\documentclass[journal]{IEEEtran}
\ifCLASSINFOpdf
\else
\fi

\usepackage{import}
\usepackage{graphicx}
\usepackage{amsmath}
\usepackage{multirow}
\usepackage{rotating}
\usepackage{romannum}
\usepackage{xcolor}
\usepackage{hyperref}
\usepackage{soul}
\hypersetup{
    colorlinks=true,
    linkcolor=blue,
    pdfpagemode=FullScreen,
    }
\date{}

\usepackage{enumitem}

\usepackage{url}

\usepackage{hyperref} 

\begin{document}

\onecolumn 

\begin{description}[labelindent=-1cm,leftmargin=1cm,style=multiline]

\item[\textbf{Citation}]{C. Zhou, G. AlRegib, A. Parchami, and K. Singh, "TrajPRed: Trajectory Prediction with Region-based Relation Learning," in \textit{IEEE Transactions on Intelligent Transportation Systems (T-ITS)}, Mar. 04, 2024.}

\item[\textbf{DOI}]{\url{https://doi.org/10.1109/TITS.2024.3381843}}

\item[\textbf{Review}]{Date of acceptance: Mar. 04, 2024\\Date of publication: Apr. 08, 2024}

\item[\textbf{Codes}]{\url{https://github.com/olivesgatech/TrajPRed}}

\item[\textbf{Bib}] {@article\{Zhou\_2024, \\
title=\{TrajPRed: Trajectory Prediction with Region-based Relation Learning\}, \\
author=\{Zhou, Chen and AlRegib, Ghassan and Parchami, Armin and Singh, Kunjan\}, \\
journal=\{IEEE Transactions on Intelligent Transportation Systems\}, \\
publisher=\{Institute of Electrical and Electronics Engineers (IEEE)\}, \\
DOI=\{10.1109/tits.2024.3381843\}, \\
url=\{http://dx.doi.org/10.1109/tits.2024.3381843\}, \\
ISSN=\{1558-0016\}, \\
year=\{2024\}, \\
pages=\{1–10\} \\
\}
}



\item[\textbf{Keywords}]{Relation Modeling, Stochastic Prediction, Trajectory Prediction, Behavior Forecasting} 

\item[\textbf{Contact}]{\href{mailto:chen.zhou@gatech.edu}{chen.zhou@gatech.edu} OR \href{mailto:alregib@gatech.edu}{alregib@gatech.edu}\\ \url{https://alregib.ece.gatech.edu/} \\ }
\end{description}

\thispagestyle{empty}
\newpage
\clearpage
\setcounter{page}{1}

\twocolumn


\title{TrajPRed: Trajectory Prediction with Region-based Relation Learning}

%

\author{Chen Zhou,~\IEEEmembership{Student Member,~IEEE,}
        Ghassan AlRegib,~\IEEEmembership{Fellow,~IEEE,}\\
        Armin Parchami,~\IEEEmembership{Snorkel AI,} 
        and~Kunjan Singh,~\IEEEmembership{Ford Motor Company} 
}

%
%

\markboth{IEEE TRANSACTIONS ON INTELLIGENT TRANSPORTATION SYSTEMS}%
{}
%

\maketitle

\begin{abstract}
Forecasting human trajectories in traffic scenes is critical for safety within mixed or fully autonomous systems. Human future trajectories are driven by two major stimuli, social interactions, and stochastic goals. Thus, reliable forecasting needs to capture these two stimuli. Edge-based relation modeling represents social interactions using pairwise correlations from precise individual states. Nevertheless, edge-based relations can be vulnerable under perturbations. To alleviate these issues, we propose a region-based relation learning paradigm that models social interactions via region-wise dynamics of joint states, i.e., the changes in the density of crowds. In particular, region-wise agent joint information is encoded within convolutional feature grids. Social relations are modeled by relating the temporal changes of local joint information from a global perspective. We show that region-based relations are less susceptible to perturbations. In order to account for the stochastic individual goals, we exploit a conditional variational autoencoder to realize multi-goal estimation and diverse future prediction. Specifically, we perform variational inference via the latent distribution, which is conditioned on the correlation between input states and associated target goals. Sampling from the latent distribution enables the framework to reliably capture the stochastic behavior in test data. We integrate multi-goal estimation and region-based relation learning to model the two stimuli, social interactions, and stochastic goals, in a prediction framework. We evaluate our framework on the ETH-UCY dataset and Stanford Drone Dataset (SDD). We show that diverse prediction benefits from region-based relation learning. The predicted intermediate location distributions better fit the ground truth when incorporating the relation module. Our framework outperforms the state-of-the-art models on SDD by $27.61\%$/$18.20\%$ of ADE/FDE metrics. Our code is available at \url{https://github.com/olivesgatech/TrajPRed}
\end{abstract}


\IEEEpeerreviewmaketitle

\section{Introduction}
\IEEEPARstart{T}{he} decision-making process within intelligent systems involves a certain cycle where perception and prediction are the primary two stages. There has been a significant advance in tackling the challenges of perception using learning-based algorithms. For instance, robust models \cite{alregib2022explanatory, lee2021open, kwon2022gating, lehman2020structures} have been developed for visual applications under challenging conditions. Additionally, the success of learning-based models on natural images has fostered its usage in computational image analysis including seismic interpretation \cite{benkert2022example, mustafa2021joint, kokilepersaud2022volumetric} and medical fields \cite{logan2022patient, logan2022multi}. While perception algorithms endow intelligent systems with the ability to perceive entities of interest, prediction of their future states suggests optimal decision-making.
Predicting the future behavior of mobile agents is essential in surveillance \cite{liang2019peeking,  ma2020autotrajectory}, sports forecasting \cite{felsen2017will, qi2020imitative}, and robotics \cite{cao2019dynamic, moller2021survey} domains. 

\begin{figure}[t]

  \centering
  {\includegraphics[width=\columnwidth]{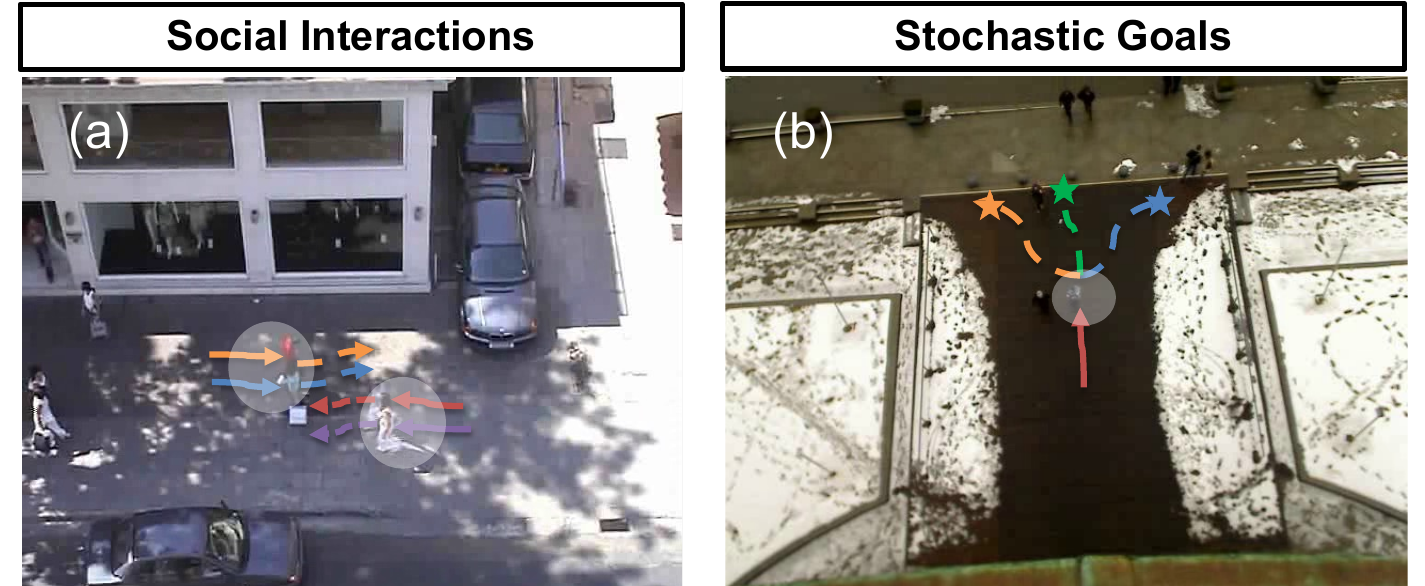}}

\caption{Example scenarios in the ETH-UCY dataset that reflect two major stimuli, social interactions and stochastic goals, in human trajectory prediction. (a) shows an example of interactions in group avoiding. Two groups of people walking toward each other avoid others along their respective future trajectories. (b) shows an example of multiple stochastic future destinations (goals). A pedestrian approaching the curbside might make a turn or follow the original direction. Plausible goals are represented by the star symbols.
Solid lines represent historical trajectories while dashed lines represent future trajectories.}
\label{fig:intro_twofactors}

\end{figure}

Human trajectory prediction is safety-critical for transportation.  In traffic prediction, human behavior is influenced by two major stimuli: social interactions and individual goals.
Consider pedestrians walking in a public area, they comply with social conventions by interacting with others. Social interactions can influence the decision-making of agents' future states. For instance, two separate groups of people walking toward each other are likely to continue walking as respective groups while avoiding others in the short-term future, as illustrated in Fig.~\ref{fig:intro_twofactors} (a). 
In addition to external social factors, human behavior is inherently goal-oriented and stochastic. In this work, we target the framework to capture the stochastic nature of goal-directed behavior in scenarios where human individuals exhibit diverse aims of reaching a particular destination. This is motivated by a number of foundational social behavior principles.
According to situated action \cite{suchman1987plans}, humans have high-level internal plans to achieve their desired destinations. Internal plans vary between individuals thereby goals can be under-deterministic. Hence, stochasticity exists in human behaviors. Given the same past observations, humans show multiple plausible future behaviors. For instance, a pedestrian walking along the sidewalk can either take a left/right turn or maintain the original direction when approaching the curbside, as illustrated in Fig.~\ref{fig:intro_twofactors} (b). Therefore, developing a prediction framework that learns social interactions and stochastic individual future goals to predict human trajectories is desirable and challenging.

\begin{figure}[t]

  \centering
  {\includegraphics[width=\columnwidth]{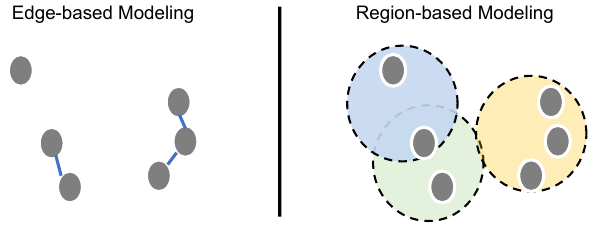}}

\caption{Two relation modeling paradigms. In an edge-based setting, the relations of individuals are modeled separately with pairwise correlations based on accurate states. In a region-based setting, relations are modeled with joint representations within each region instead of accurate individual states.}
\label{fig:interaction_model_granularity}

\end{figure}

Several approaches have attempted to model the aforementioned stimuli. The authors in \cite{alahi2016social, salzmann2020trajectron++, gupta2018social} incorporate pairwise human-human interactions. These interaction modeling methods are edge-based that encode behavior correlations from pairs of agents, as illustrated in Fig.~\ref{fig:interaction_model_granularity} (left). Edge-based modeling methods can be less adaptable to different scenes. They restrict social interaction to predefined neighborhoods that vary across scenes with different structures and densities. Furthermore, edge-based methods can be less robust to perturbations on position states since they construct interaction representations directly from individual locations. To alleviate these issues, we learn region-based relations. At a high level, local interactions of agents within a region of interest are modeled in a joint manner, as illustrated in Fig.~\ref{fig:interaction_model_granularity} (right). Specifically, informative local interactions are represented by dynamics within regions. Instead of modeling social interactions from pairwise agent correlations, we utilize region-wise dynamics as the basis. Individuals are influenced by the dynamics of the joint information, i.e., how the density of crowds changes, in each region. While edge-based modeling relies on precise spatial position states, region-based modeling constructs relations from smoothed density thereby being less susceptible to perturbations on positions. We propose to relate region-wise temporal dynamics of agents' joint states to learn how social relations evolve from one vicinity to another from a global perspective. In particular, the joint spatial states of all agents within a vicinity are encoded in the grids of latent maps of fully-convolutional networks. The inter-regional relation representations are then characterized using the temporal changes of these feature grids to model social relations. In this way, our framework constructs relation representations using region-wise correlations instead of edge-based human-human correlations, leading to enhanced robustness under the perturbation in individual spatial states.

Besides social interactions, individual goals can also influence future behavior. We estimate the agent's goal and use the estimated goal as part of the inputs to predict future states. 
Individual goals can be under-deterministic since they are influenced by internal decisions. Empirically, we find the discrepancy in displacement errors between training and test data when conditioning prediction on deterministic goals. The goal learned from training data does not fully reflect the behavior in the test scenarios, motivating stochastic goal estimation. To account for human behavior stochasticity, we adopt conditional variational autoencoder (CVAE), a generative technique inspired by multi-modal prediction methods\cite{salzmann2020trajectron++, mangalam2020not}, to realize stochastic multi-goal estimation in future prediction. 

We propose a prediction framework that combines region-based relation learning and stochastic goal estimation to model the two stimuli. The proposed framework possesses several advantages. We show that the generated prediction distributions from our framework with region-based relation learning better fit the future locations. Additionally, we find that our method, compared to edge-based relation learning approaches, is more robust to the perturbations of the agent's positions. Furthermore, we compare our method with state-of-the-art methods on the ETH-UCY dataset and Stanford Drone Dataset. We provide an analysis of the behavior of several deterministic and generative models and show the efficacy of conditioning the prediction on stochastic goal estimation. The results demonstrate that our framework performs superior or comparable to several state-of-the-art approaches.

In summary, the contributions of our framework are three-fold:
\begin{enumerate}
    \item We propose a robust relation learning paradigm via region-wise temporal dynamics to model the social compliance between humans. We find that our framework is more robust to spatial noise perturbations compared to edge-based relation learning approaches.
    \item We estimate multiple plausible goals that account for human stochasticity by adopting CVAE, a generative technique inspired by multi-modal prediction. We show the efficacy of conditioning prediction on stochastic goal estimation.
    \item We integrate the proposed region-based relation learning module and the multi-goal estimation into a framework to model two major stimuli: social compliance and individual goals. We show that the diverse prediction benefits from the region-based relation module. 
\end{enumerate}




\section{Related Work}
In this section, we review related research and discuss the distinctions and advantages of the proposed prediction framework. Generally, there have been many prior works \cite{rudenko2020human} studying human future prediction. Several early works \cite{keller2013will, goldhammer2014pedestrian} extract motion patterns from the observed data to predict future trajectories. Many later works incorporate social interactions between humans that influence the future trajectory. Recent works have developed generative approaches to account for the stochasticity of trajectory prediction by generating a distribution of plausible future trajectories given past observations. Additionally, there have been works on goal-directed path planning, which consider the agent’s goal while predicting the future path.

\begin{figure}[t]

  \centering
  {\includegraphics[width=\columnwidth]{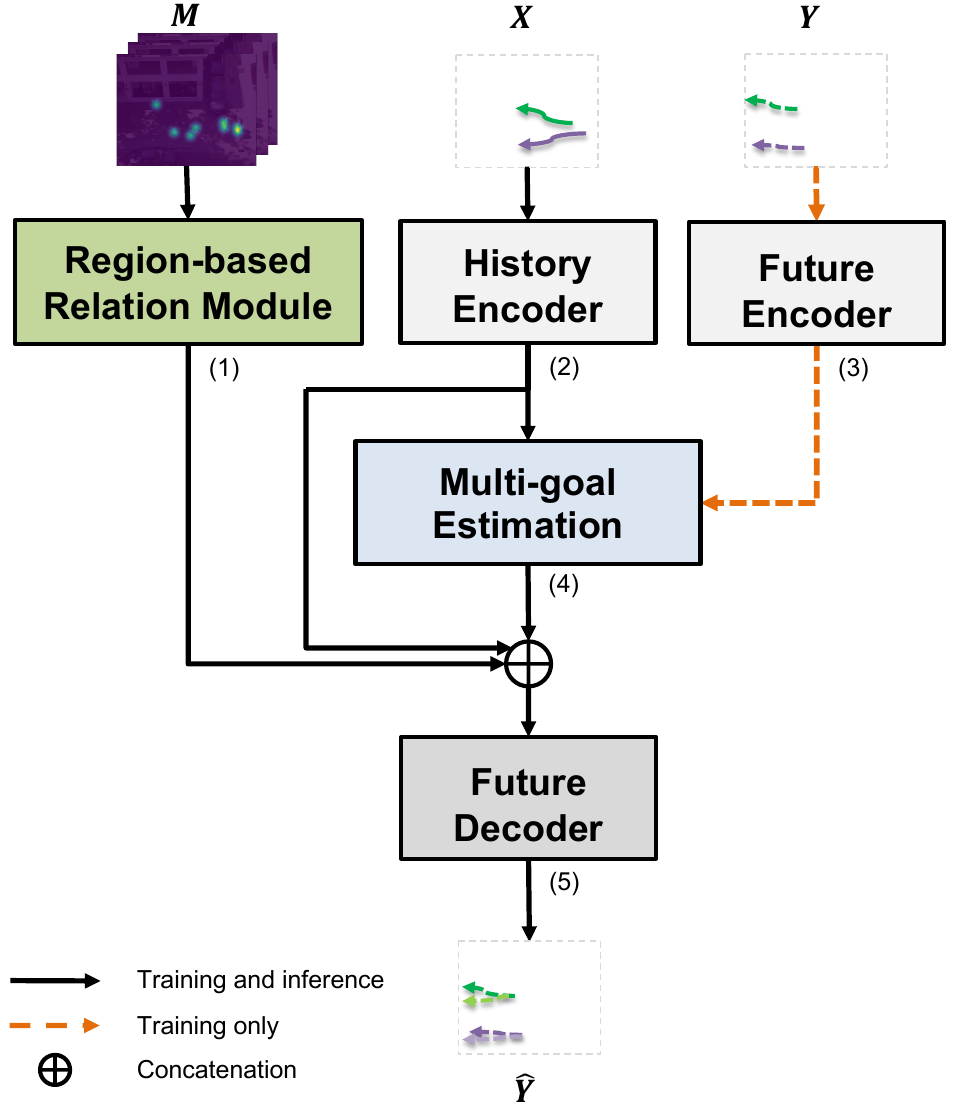}}

\caption{{The overall diagram of the proposed trajectory prediction framework.}  The region-based joint relations are learned via the Relation Module using the trajectory maps $M$. The History Encoder and the Future Encoder encode the individual observed and the future trajectory patterns, $X$, $Y$, respectively. The Multi-goal Estimation generates multiple end positions conditioned on the encoded features from the history and future encoders. The Future Decoder predicts the future trajectories $\hat{Y}$ using the combination of joint relation features, individual history trajectory features, and estimated end positions. The orange dashed arrows, and the black arrows stand for the data flow in training only, and both training and inference, respectively.}
\label{fig:overall_diagram}
\end{figure}

\textbf{Interaction-aware Prediction}
Mobile agents comply with social conventions by interacting with others. Traditional approaches utilize hand-crafted rules to model social interactions. For instance, social force models \cite{helbing1995social, mehran2009abnormal, yamaguchi2011you, alahi2014socially} capture pedestrian interactions as attractive and repulsive forces to avoid collisions and achieve their goals. These approaches are limited to representing high-level social behavior patterns, especially when data become massive. Hence, data-driven approaches have become increasingly prevalent to predict future trajectories. 
In particular, existing interaction-aware approaches use recurrent neural networks (RNNs) to predict human trajectory \cite{bartoli2018context, manh2018scene}. In general, these methods utilize a recurrent encoder-decoder backbone where a recurrent encoder encodes individual agent information and interactions from past observation data, and a recurrent decoder generates future states recurrently utilizing information extracted from observation data. For instance, Social LSTM \cite{alahi2016social} models individual agent's states with LSTM and models interactions between spatially proximal agents through a pooling mechanism. Similar to Social LSTM, Social Attention \cite{vemula2018social} weights interactions between all agents via an attention mechanism. These interaction modeling approaches can be viewed as edge-based since they model agent-agent influences via independent pair-wise correlations between agent spatial states. Instead of representing social relations from individual positions, our method learns relations via region-wise dynamics of density that represent the changes in the joint agent state information. 

\textbf{Multi-future Generative Approaches}
The aforementioned works focus on deterministic predictions. However, given the same observation data, an agent is likely to take multiple plausible future paths depending on their short-term goals. Hence, other algorithms aim to model the stochasticity of agents to perform multi-future prediction \cite{liang2020garden, li2021spatio, li2019conditional}. Some algorithms model stochasticity via conditional variational autoencoder (CVAE) \cite{kingma2013auto}. For instance, Trajectron++ \cite{salzmann2020trajectron++} represents trajectories as dynamic spatiotemporal graphs, and uses variational generative modeling to generate distributions of future trajectories. Another line of work uses generative adversarial training \cite{NIPS2014_5ca3e9b1}. For instance, Social GAN \cite{gupta2018social} uses a GAN-LSTM architecture to generate multiple socially-plausible future predictions. SoPhie \cite{sadeghian2019sophie} further incorporates attention modules to model agents’ interactions with the scene environment and other agents. While deterministic approaches are evaluated on the deterministic model outcomes, during the inference,  generative approaches are evaluated on their optimal outcomes from the distribution using the associated ground truth. Hence, in contrast to deterministic approaches that learn average behavior, generative approaches learn desirable behavior, i.e., the optimal prediction among multiple plausible predictions.

\section{Technical Approach}




In this section, we define the problem of trajectory prediction and describe the region-based relation module and multi-goal estimation for learning social relations and the stochasticity of mobile agents. We first formulate the prediction problem in subsection \hyperref[subsec:problemdefine]{III-A}. We describe the framework overview in subsection \hyperref[subsec:sysoverview]{III-B}. Subsection \hyperref[subsec:regionencoding]{III-C} and \hyperref[subsec:regionrelation]{III-D} present our region-based relation learning workflow. Subsection \hyperref[subsec:stochgoalest]{III-E} describes the multi-goal estimation module. The overall training objective is described in subsection \hyperref[subsec:FFT]{III-F}. We present the contributions and technical details of each key module of the proposed framework in the following.

\subsection{Problem Definition} \label{subsec:problemdefine}
In trajectory prediction, we aim to forecast the future states of mobile agents given their history states. Let $X_i = [X_i^{t-\tau+1},..., X_i^{t}]$ denote the observed history states of $i^{th}$ agent from time $t-\tau+1$ to $t$, and ${Y}_i=[{Y}_i^{t+1},...,{Y}_i^{t+T}]$ denote the associated future states from $t+1$ to $t+T$. The goal of prediction is to learn a mapping from the input $X_i$ to the predicted output $\hat{Y}_i$. In this work, the input $X_i^{t}$ takes the form of observed center positions, estimated velocities, and accelerations in a bird's-eye-view. The output $\hat{Y}_i^{t+T}$ consists of future center positions. Without loss of generality, we assume that the positions in $X_i$ are obtained from an object detection and tracking framework in advance.

\begin{figure*}[th]

  \centering
  {\includegraphics[scale=0.4]{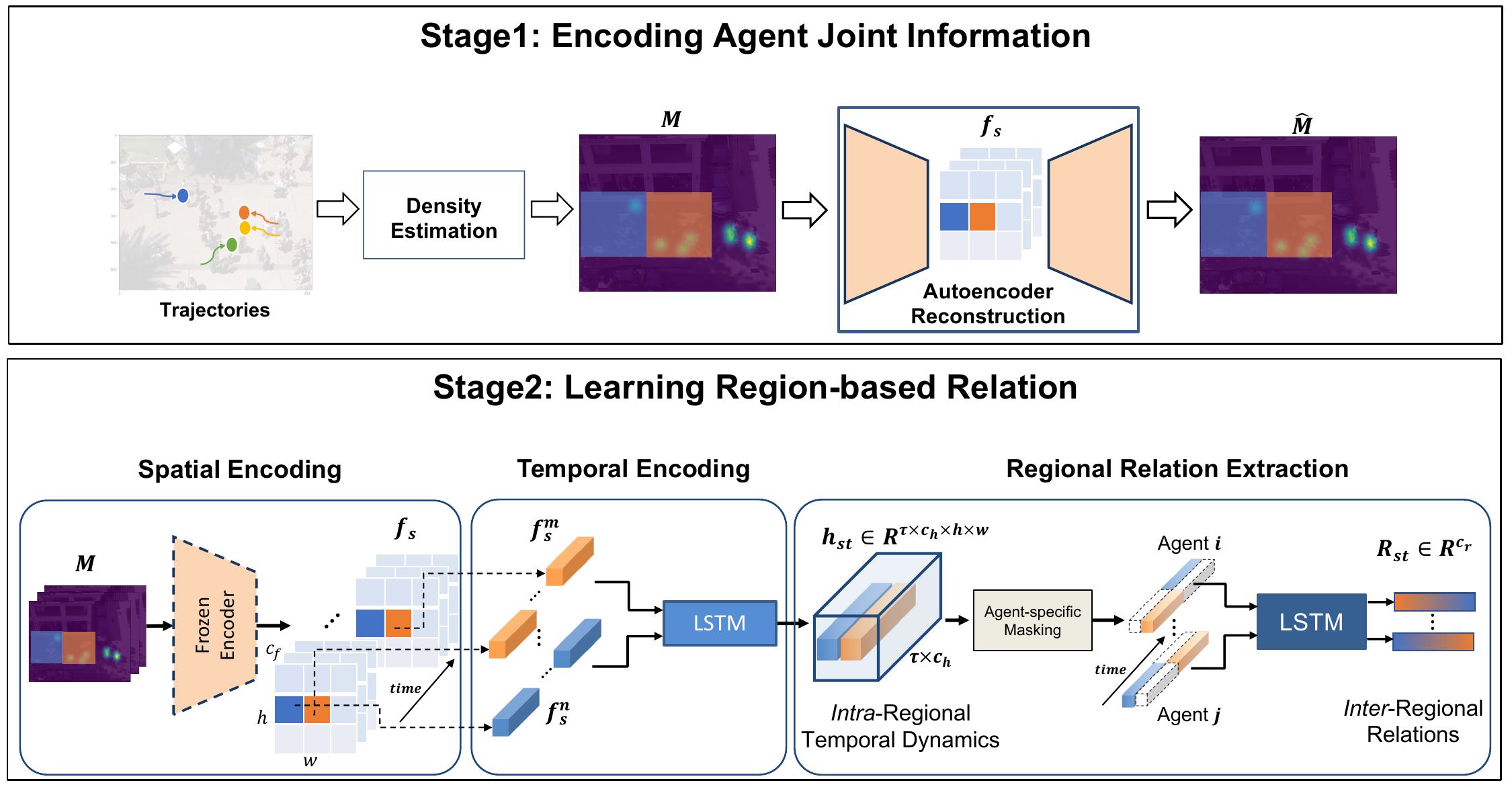}} 

\caption{The overall two-stage workflow of the proposed regional relation module. In the first stage, region-wise agent joint information is encoded within the grids of the latent feature maps $f_s$ via auto-encoding. In the second stage, the \textit{intra}-regional interaction features $h_{st}$ are extracted via Temporal Encoding using the latent feature grids $f_s^m,...,f_s^n$ obtained by the pre-trained autoencoder. The \textit{inter}-regional relation representations $R_{st}$ are learned via relating the agent-specific $h_{st}$ evolving across regions in the Regional Relation Extraction.}
\label{fig:TrajPRed_relation_module}
\end{figure*}

\subsection{Framework Overview}
\label{subsec:sysoverview}
The high-level diagram of our framework is illustrated in Fig.~\ref{fig:overall_diagram}. Our framework consists of five major components: (1) a relation module that learns agent joint social relations via the region-based temporal dynamics; (2) a history encoder that encodes the observed individual trajectory patterns; (3) a future encoder extracting the future individual trajectory patterns during training; (4) a multi-goal estimation block that accounts for the stochasticity of human's short-term future; (5) a future decoder that predicts future trajectories using the combination of the features generated by the aforementioned blocks. We present detailed descriptions of these components in the following subsections.

\subsection{Encoding Region-wise Agent Joint Information} 
\label{subsec:regionencoding}
We encode region-based agent joint spatial information to account for social interactions. To realize this, we employ a regular convolutional autoencoder that reconstructs the trajectory maps as illustrated in the first stage in Fig.~\ref{fig:TrajPRed_relation_module}. Our intuition is that the grids of the autoencoder latent feature maps encode the regional information associated with the spatial patches in the scene. We obtain the density of crowds to represent regional information and encode changes in density as regional dynamics. Specifically, we generate trajectory maps $M$ by applying Gaussian filtering on the ground-truth trajectories in each scene frame. As illustrated in Fig.~\ref{fig:TrajPRed_relation_module}, the convolutional autoencoder encodes the receptive fields, e.g., shaded blue and orange patches, within the input trajectory maps $M$ into a set of associated latent feature vectors $f_s^1,...f_s^{h\times w}$, where ${h\times w}$ denotes the spatial dimension of the latent feature maps. The latent features are then passed through the decoder to generate reconstructed maps $\hat{M}$.

\subsection{Learning Region-based Relation Representations}
\label{subsec:regionrelation}
The agents interact within their local neighborhood, which is reflected by the dynamics within each region over time. Relating the dynamics of local interactions across different regions informs how agents' social behaviors change from one region to another from a global perspective, which motivates our region-based relation learning in the second stage in Fig.~\ref{fig:TrajPRed_relation_module}. 

Given the autoencoder pre-trained for trajectory map reconstruction, we utilize the temporal dynamics of its latent features to learn regional relations. The observed trajectory maps $M=[M^{t-\tau+1},...,M^{t}]$ are passed through the encoder part of the autoencoder to obtain a set of frame-level features $f_s=[f_s^1,...f_s^{h\times w}]\in{\rm I\!R}^{\tau\times c_f\times h\times w}$, where $f_s^{k}\in {\rm I\!R}^{\tau\times c_f}$ represents the agent joint spatial distribution in $k$-th region from ${t-\tau+1}$ to ${t}$. Here, $c_f$, $h$, and $w$ denote the channel dimension, height, and width of the feature maps, respectively. The feature maps $f_s$ are then passed through a temporal encoding module to generate spatiotemporal local interaction representations $h_{st}=[h_{st}^{1},..,h_{st}^{k},...,h_{st}^{h\times w}]\in{\rm I\!R}^{\tau\times h\times w\times c_h}$. Here, $c_h$ denotes the channel dimension of the local interaction representations. Specifically, we employ an LSTM as temporal encoding to extract the temporal dynamics $h_{st}^{k}\in {\rm I\!R}^{\tau\times c_h}$ from each sequence of feature grid $f_s^{k}$. Here, $h_{st}^{k}$ denotes the collected hidden states of the LSTM from ${t-\tau+1}$ to ${t}$ within each region, which can be viewed as {intra}-regional interaction features capturing the region-wise sequence’s historical information up to their associated time step. The intra-regional interaction dynamics change between different local regions when the agent moves around in the scene. To learn the agent's social interactions from a global perspective, the intra-regional temporal dynamics $h_{st}$ are passed through a regional relation extraction module, consisting of an LSTM, to generate the relation features $R_{st}\in {\rm I\!R}^{c_r}$. Here, $c_r$ denotes the channel dimension of the final relation representations. The intra-regional features $h_{st}$ are agent-agnostic representations. Given the same observed trajectory maps $M$, the behavior of different agents can be influenced by different intra-regional temporal dynamics when moving across different regions. To learn the distinct relation context for each target agent, we employ an agent-specific masking mechanism before passing $h_{st}$ to the regional relation extraction module. The agent-specific masking selects the sequence of temporal dynamics $h_{st}$ according to the regions that the agent traversed. For instance, the masked intra-regional temporal dynamics for $i$-th agent is represented as $[\{h_{st}^{n}\}^{t-\tau+1},...,\{h_{st}^{m}\}^{t_o}...,\{h_{st}^{m}\}^t]$, when the agent moves across from the $n$-th region to the $m$-th region at time $t_o$. For every agent in the observed trajectory maps, the LSTM in the regional relation extraction module relates the masked intra-regional features to generate the \textit{inter}-regional relation representations $R_{st}$. The relation representations $R_{st}$ are utilized later to predict future trajectories. 
{Finally, note that GRU can be used as the backbone of the Region-based Relation Module as well.}
\begin{figure}[t]

  \centering
  {\includegraphics[width=\columnwidth]{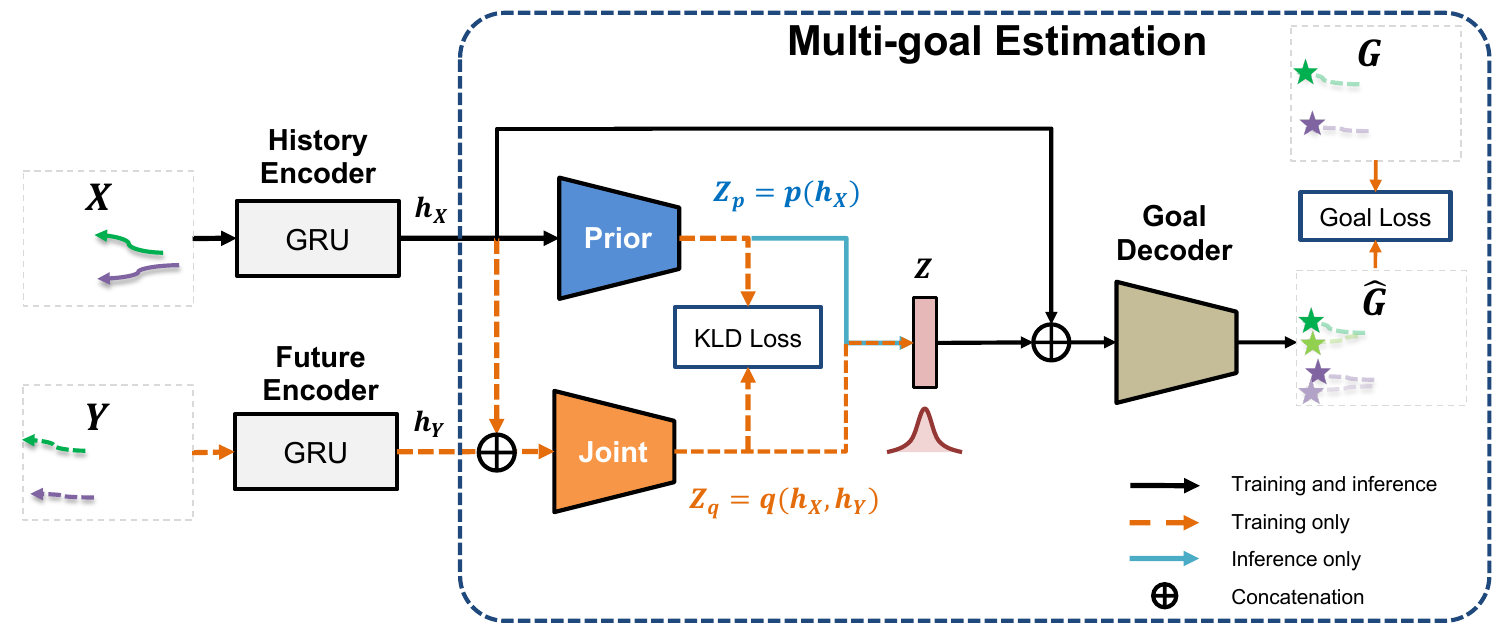}}

\caption{The workflow of stochastic goal estimation. The orange dashed arrows, the blue arrows, and the black arrows stand for the data flow in training only, inference only, and both training and inference, respectively.}
\label{fig:TrajPRed_SGE}
\end{figure}

\subsection{Estimating Stochastic Goals}
\label{subsec:stochgoalest}
In addition to modeling the external social relations, we estimate multiple plausible goals (future end positions) to account for the stochasticity of agents. Specifically, we employ a CVAE, of which the latent representations are drawn from Gaussian variables, within a multi-goal estimation workflow as illustrated in Fig.~\ref{fig:TrajPRed_SGE}. The outcome goals are decoded using the latent representations sampled from the Gaussian distribution. In contrast to the methods that assume a specific prediction distribution, e.g., Gaussian Mixture Model (GMM), of the CVAE \cite{ivanovic2019trajectron, salzmann2020trajectron++}, we do not impose parametric distributions on the stochastic goal estimation. 

Overall, the multi-goal estimation module takes the hidden representations of the individual history and future trajectories as inputs to decode the future end positions $\hat{Y}_i^{t+T}$ of each agent independently. Given the agent's individual history trajectory $X$ and future trajectory $Y$ during training, the history encoder and the future encoder encode $X$ and $Y$ into the hidden representations $h_{X}$ and $h_{Y}$, respectively. $h_{X}$ and $h_{Y}$ summarize the past and future agent-specific motion patterns. Both the history encoder and the future encoder consist of a gated recurrent union (GRU). Note that we assume the future trajectory $Y$ is available during training. The X-Y joint network $q$ (orange) projects the concatenated representations from $h_{X}$ and $h_{Y}$ to the latent distribution representation $Z_q$. $Z_q=q(h_{X},h_{Y})$ captures the dependency between the history $X$ and future $Y$. Meanwhile, the prior network $p$ (blue) projects $h_{X}$ to the latent distribution representation $Z_p$. Kullback–Leibler divergence (KLD) loss between $Z_p$ and $Z_q$ is optimized so that the prior network $p$ implicitly learns the dependency between $h_Y$ and $h_X$. Consequently, prior $p$ captures the dependency between individual future goals and past states from only input history $X$. However, the individual goals learned by a deterministic goal estimation from the training data do not fully reflect human behavior in the test data due to stochasticity. Hence, we introduce a non-deterministic sampling to estimate stochastic goals. During training, the latent representation $Z$ is sampled from a Gaussian distribution parameterized by $Z_q$. The goal decoder network projects the concatenated representations from $Z_q$ and $h_X$ to multiple estimated goals $\hat{G}$. We use a mean-squared error (MSE) criteria to penalize the deviation between the estimated goal $\hat{G}$ and the ground-truth goal $G$. During inference, the latent representation $Z$ is sampled from a Gaussian distribution parameterized by $Z_p$ via only the prior network $p$. $Z$ is concatenated with $h_X$ to estimate $\hat{G}$. Here, we use multi-layer perceptrons (MLPs) for the prior, joint, and goal decoder networks.

\subsection{Forecasting Future Trajectories}
\label{subsec:FFT}
The social relations, individual stochastic goals in combination with  past motion patterns together influence the decision-making of a target agent and the associated future states. Hence, we utilize the outcomes of the aforementioned relation module, the multi-goal estimation module to regress the agent's future trajectories. Specifically, the future decoder in Fig.~\ref{fig:overall_diagram}, consists of a GRU, takes the concatenated representations from $R_{st}$, $\hat{G}$ and $h_X$ to forecast the future trajectory $\hat{Y}=[\hat{Y}^{t+1},...,\hat{Y}^{t+T}]$. We employ an MSE objective to penalize the deviation between predicted trajectory $\hat{Y}$ and the ground-truth trajectory $Y$ at each time step along the prediction horizon. 

To train the full proposed framework, we employ the following losses:
\begin{equation}
\begin{aligned}
	\mathcal{L} = & \mathcal{L}_{goal} + \mathcal{L}_{traj} + \mathcal{L}_{KLD} \\
	= & \min_{k\in K} (\|G-\hat{G}^k\|_2  + \|Y-\hat{Y}^k\|_2) + \beta KL(Z_q||Z_p)
\end{aligned}
\end{equation}
where $K$ denotes the number of sampled future goals, and $\beta$ denotes the weight coefficient for KL divergence. Note that we generate $K$ goals $\hat{G}^k, k=1,...,K$, the estimated $\hat{G}^k$ closest to the ground-truth $G$ is selected to compute the goal loss $\mathcal{L}_{goal}$. Conditioned on this closest $\hat{G}^k$, the associated future trajectory $\hat{Y}^k$ is used to calculate the trajectory loss $\mathcal{L}_{traj}$.


\section{Experiments}
In this section, we empirically evaluate our prediction framework on the ETH\cite{pellegrini2009you}-UCY\cite{lerner2007crowds} dataset, {and Stanford Drone Dataset (SDD)}\cite{robicquet2016learning}. They contain pedestrian trajectories in urban scenes. We first present the experimental setting and evaluation metrics. We validate the effectiveness of the major components through baseline experiments. We then discuss the comparison between our method and several related approaches.
\subsection{Dataset and Preprocessing}
\textbf{Dataset}
The ETH-UCY dataset consists of ETH, Hotel, Univ, Zara1, and Zara2 outdoor scenes captured from a bird’s-eye view. All scenes contain human trajectories in world coordinates (in meters) annotated at $2.5$ FPS. These scenes contain various human activities, e.g., person following, groups crossing each other, groups merging and dispersing, etc. {Stanford Drone Dataset (SDD) is a large-scale dataset that consists of  $60$ aerial-view videos captured by drones over Stanford University. SDD contains positions of more than $11,000$  traffic agents with various  types, e.g., pedestrians, bicycles, and cars, annotated with bounding boxes in pixel coordinates.}

\textbf{Preprocessing Data}
To extract social relations, we use the ground-truth trajectories of each scene to construct trajectory density maps that represent the joint states of all agents in the scene. Trajectory density maps can represent crowd behavior from noisy measurement data, in which the errors can be carried from tasks prior to prediction. Specifically, we create frame-wise trajectory density maps using a two-dimensional Gaussian filter. 

\subsection{Experimental Setting}
\textbf{Evaluation Schemes} For the ETH-UCY dataset, we follow the commonly used leave-one-out evaluation protocol as previous works \cite{alahi2016social, vemula2018social, gupta2018social, sadeghian2019sophie, salzmann2020trajectron++} i.e., training on four scenes and evaluating on the fifth held-out scene. {For Stanford Drone Dataset (SDD), we follow the standard data splitting used in} \cite{su2022trajectory, xia2022cscnet} {including $36$ training videos, $12$ validation videos, and $12$ test videos.} The training and test data consist of sequences with a fixed duration of $8$ seconds, where the first $\tau=3.2$ seconds ($8$ frames) of observation sub-sequences are inputs, and the next $T=4.8$ seconds ($12$ frames) of future sub-sequences are prediction ground truth. The sequences are sampled with a stride of 1 frame. We assume the future sub-sequences are only available during the training.

\textbf{Evaluation Metrics} Following prior works, our prediction framework is evaluated based on the following metrics:
\begin{enumerate}
    \item \textit{Average Displacement Error (ADE)}: average $L_2$ distance between the ground truth and predicted future trajectories, over all agents and all time steps.
    \item \textit{Final Displacement Error (FDE)}: $L_2$ distance between only the predicted final location and the ground truth final location at the end step of the prediction horizon.
    \item \textit{Kernel Density Estimate based Negative Log-Likelihood (KDE NLL)}: average NLL of the ground-truth trajectory under a stochastic prediction distribution constructed by fitting a kernel density estimation on generated trajectory samples \cite{thiede2019analyzing}.
\end{enumerate}
\subsection{Baseline and Ablation Analysis}
In order to develop an understanding of the influence of the proposed region-based relation module and the multi-goal estimation module, we conduct the following baseline and ablation study. We evaluate our framework without these two modules, i.e., the baseline model in Table~\ref{Tab: baseline_and_ablation}. The baseline model is compared with the framework that integrates only the region-based relation module, and the framework that integrates only the multi-goal estimation module.

\textbf{Learning Social Interactions via Region-based Relation Modeling} To study the impact of the region-based relation learning module, we compare the ADE/FDE of the predictions of the baseline model and the framework with only the relation module (Baseline$+$R)
 as shown in Table~\ref{Tab: baseline_and_ablation}. The framework with region-based relation learning shows lower ADE/FDE in all data scenes compared to the baseline, demonstrating that modeling the region-based social interactions improves the location prediction baseline.

\def\arraystretch{1.5}%
\begin{table}[t]
\caption{{Ablation study (ADE/FDE in meters) results of the influence of the relation module and the goal estimation module on the ETH-UCY datasets. Lower is better. Baseline$+$R: add only the region-based relation module to the baseline. Baseline$+$G: add only the multi-goal estimation module to the baseline. Baseline$+$R$+$G: add the two proposed modules to the baseline.} }

  \begin{center}
 \setlength{\tabcolsep}{3pt}
\begin{footnotesize}

\begin{tabular}{c|c|c|c|c}
\hline
\multirow{ 2}{*}{Dataset} & \multicolumn{2}{c|}{ADE/FDE} & \multicolumn{2}{c}{ADE/FDE (Best of $20$ Predictions)}             \\ 
 & \multicolumn{1}{c|}{Baseline} &{Baseline$+$R} & {Baseline$+$G}  & \begin{tabular}{@{}c@{}}Baseline$+$R$+$G\\ (TrajPRed) \end{tabular} \\ \hline
ETH  &      1.21/2.42 &  1.10/2.07 &  0.52/0.83 &   0.54/0.87 \\ \hline
Hotel &     0.44/0.90 &  0.29/0.51 & 0.15/0.23 &  0.14/0.22 \\  \hline
Univ &      0.73/1.58 &  0.68/1.44 & 0.30/0.55 &   0.28/0.54 \\ \hline
Zara 1 &    0.52/1.15 &  0.39/0.84 & 0.21/0.36 &    0.20/0.36 \\ \hline
Zara 2 &    0.43/0.94 &  0.31/0.66 & 0.17/0.30 &  0.16/0.29 \\ \hline

\end{tabular}
    
\end{footnotesize}
\end{center}

  \label{Tab: baseline_and_ablation}
\end{table}

\def\arraystretch{1.5}%

\textbf{Capturing Human Stochasticity via Non-deterministic Goal Estimation}
We then compare the predictions of the baseline and the framework with only the goal estimation (Baseline$+$G in Table~\ref{Tab: baseline_and_ablation}) to assess the influence of the multi-goal estimation module.
Here, the ADE/FDE results of the model Baseline$+$G are calculated based on the optimal outcome of the predictions generated from $20$ estimated goals. We observe the discrepancy in FDE and ADE between the method with and without stochastic sampling for goal estimation. The large margin indicates that the behavior learned by the deterministic baseline from the training data does not fully reflect the behavior in the test data. This discrepancy is caused by the stochasticity of human behavior across different unconstrained environments, demonstrating the importance of incorporating multi-goal estimation.

\def\arraystretch{1.5}%
\begin{table*}[th]
\centering
\caption{Trajectory prediction results (ADE/FDE in meters) of baselines and our approach on the ETH-UCY datasets. Lower is better. Generally, generative models encourage a diverse set of plausible predictions capturing human stochasticity. Hence, their optimal predictions are closer to the ground truth than the predictions of deterministic methods.}
\setlength{\tabcolsep}{3pt}
\begin{scriptsize}
\begin{tabular}{c||llllllllll|l}
\hline
\multicolumn{1}{l||}{Dataset} & \multicolumn{1}{c}{Linear} & \multicolumn{1}{c}{LSTM} & \multicolumn{1}{c}{S-LSTM\cite{alahi2016social}} & \multicolumn{1}{c}{S-GAN\cite{gupta2018social}} & \multicolumn{1}{c}{SoPhie\cite{sadeghian2019sophie}} & \multicolumn{1}{c}{PECNet\cite{mangalam2020not}} & {Trajectron} \cite{ivanovic2019trajectron} & \multicolumn{1}{c}{Trajectron++\cite{salzmann2020trajectron++}}  & \multicolumn{1}{c}{{CSCNet}\cite{xia2022cscnet}} & \multicolumn{1}{c|}{{VDRGCN}\cite{su2022trajectory}} & \multicolumn{1}{c}{TrajPRed} \\ \hline
ETH                          & 1.33/2.94                  & 1.09/2.41                & 1.09/2.35                  & 0.81/1.52                 & 0.70/1.43                  & \textbf{0.54}/{0.87}     & 0.59/1.14             & 0.67/1.18       & 0.51/1.05   & 0.62/\textbf{0.81}               & \textbf{0.54}/{0.87}                   \\ \hline
Hotel                        & 0.39/0.72                  & 0.86/1.91                & 0.79/1.76                  & 0.72/1.61                 & 0.76/1.67                  & 0.18/0.24        & 0.35/0.66          & 0.18/0.28     & 0.22/0.42    & 0.27/0.37               & \textbf{0.14}/\textbf{0.22}                    \\ \hline
Univ                         & 0.82/1.59                  & 0.61/1.31                & 0.67/1.40                  & 0.60/1.26                 & 0.54/1.24                  & 0.35/0.60      & 0.54/1.13            & 0.30/\textbf{0.54}    & 0.47/1.02   &  0.38/0.58                & \textbf{0.28}/\textbf{0.54}                    \\ \hline
Zara 1                       & 0.62/1.21                  & 0.41/0.88                & 0.47/1.00                  & 0.34/0.69                 & 0.30/0.63                  & 0.22/0.39        & 0.43/0.83          & 0.25/0.41   & 0.36/0.81      & 0.29/0.42               & \textbf{0.20}/\textbf{0.36}                    \\ \hline
Zara 2                       & 0.77/1.48                  & 0.52/1.11                & 0.56/1.17                  & 0.42/0.84                 & 0.38/0.78                  & 0.17/0.30       &  0.43/0.85           & 0.18/0.32   & 0.31/0.68     &  0.21/0.32               & \textbf{0.16}/\textbf{0.29}                    \\ \hline
\multicolumn{1}{l||}{Average} & 0.79/1.59                  & 0.70/1.52                & 0.72/1.54                  & 0.58/1.18                 & 0.54/1.15                  & 0.29/0.48        &  0.47/0.92          & 0.32/0.55    & 0.37/0.79        & 0.35/0.50           & \textbf{0.27}/\textbf{0.46}                    \\ \hline
\end{tabular}
\end{scriptsize}
\label{Tab:comparative_results_ethucy}
\end{table*}

\subsection{Comparison with State-of-the-art Frameworks}
In addition to the baseline and ablation study, we compare our full model, TrajPRed, with the following several state-of-the-art prediction frameworks and further discuss our contributions.
\begin{enumerate}
    \item \textit{Linear}: a linear regression model that estimates parameters by minimizing the least square error.
    \item \textit{LSTM}: a vanilla LSTM model leveraging only the agent history information.
    \item \textit{S-LSTM} \cite{alahi2016social}: an LSTM-based encoder-decoder model. The latent features of spatially proximal agents are pooled to model social interactions.
    \item \textit{S-GAN} \cite{gupta2018social}: in contrast to deterministic methods, \textit{S-GAN} introduces a generative adversarial learning paradigm based on \textit{S-LSTM} to generate plausible trajectory outcomes.
    \item \textit{SoPhie} \cite{sadeghian2019sophie}: based on LSTM-GAN, the model employs both scene and social attention mechanisms to account for the scene as well as social compliance.
    \item \textit{PECNet} \cite{mangalam2020not}: a stochastic goal-conditioned method that incorporates social influence from neighboring agents.
    \item \textit{Trajectron++} \cite{salzmann2020trajectron++}: a graph-structured generative model with dynamic constraints to predict distributions of dynamically-feasible future trajectories.
    \item {\textit{CSCNet}} \cite{xia2022cscnet}: {a context-aware transfer model that alleviates the discrepancies of physical and social interactions across scenes.}
    \item {\textit{VDRGCN}} \cite{su2022trajectory}: {a graph convolutional network-based model that encodes three directed graph topologies to exploit different types of social interactions.}
\end{enumerate}

We first compare our method against the aforementioned frameworks on the ADE and FDE metrics and summarize the experimental results of these approaches on ETH-UCY in Table ~\ref{Tab:comparative_results_ethucy}.

\textbf{Behavior of Deterministic Models} The deterministic approaches learn a mapping directly from the observation input $X$ to the future deterministic output $Y$. The predictions of the deterministic approaches, e.g., Linear, LSTM, S-LSTM, generally result in a larger deviation from the ground truth with the overall higher ADE/FDE in Table ~\ref{Tab:comparative_results_ethucy}. Among the deterministic approaches, the Linear model performs the worst since it is not capable of modeling long-term temporal dependency and does not incorporate complex social interactions. {S-LSTM} improves over the {Linear} model, since it utilizes a recurrent backbone to capture temporal dependency, and employs a social pooling mechanism to model interactions.
{CSCNet further alleviates the semantic discrepancy of social and physical interactions across various scenes and scenarios to enhance prediction accuracy.}

\textbf{Behavior of Generative Models} In contrast to the deterministic algorithms, the generative approaches encourage a diverse set of plausible predictions by learning one-to-many mappings. Common evaluation metrics for these approaches include the minimum (optimal) ADE and FDE from all $K$ generated predictions \cite{salzmann2020trajectron++, mangalam2020not}. During inference, some generative methods (e.g., {S-GAN}, {SoPhie}, {Trajectron++}) generate all $K=20$ trajectories and then select the prediction closest to the ground truth with minimum ADE for evaluation. In addition to selecting the candidate trajectory based on the minimum ADE, the goal-based methods (e.g., {PECNet} and {TrajPRed}) select the candidate prediction based on the minimum FDE. Specifically, these methods generate multiple (e.g., $K=20$) destinations as goals, and the goal closest to the ground-truth endpoint is then selected. To evaluate the trajectory, the future positions are generated by conditioning on this optimal goal. Compared to other methods, {PECNet} and TrajPRed better estimate the agent's goal as their FDE results are lower (0.48 and 0.46 respectively) since they both explicitly minimize the lowest error of goal estimation during training. While {PECNet} estimates the plausible goals using the future endpoint and the past trajectory, TrajPRed utilizes the future trajectory and the past trajectory to estimate goals. Leveraging the patterns of the full future trajectory better characterizes the pedestrians' intent, leading to overall lower destination displacement errors.
{In addition to the overall performance, we find that our method outperforms PECNet in all scenes except the ETH scene. The scene-wise performance discrepancy can be attributed to the difference in the level of social interactions among various scenes. Compared to other scenes, the pedestrians in the ETH scene are less interactive with more uncertain future positions. 
PECNet and our method achieve the same prediction performance on ETH, since they both condition the stochastic sampling on future destinations to generate diverse trajectories. For other scenes, our method achieves lower prediction errors since the region-based relation module better captures the social interactions in those scenarios.}
{We also observe our method's performance decline in FDE compared to VDRGCN in ETH. The underlying reason is that VDRGCN utilizes the fat tail property of Cauchy distributions to generate more diverse predictions. Thus, it is effective especially in the ETH scene that contains more trajectories with uncertainty in future positions.}

\begin{table}[th]
\caption{Average NLL results of our framework and other approaches on ETH-UCY datasets. Lower is better. Conditioning on the estimated goal generates distributions that better fit the human behavior of the test data. TrajPRed w/o R: the model with only the multi-goal estimation module.}

  \begin{center}
 \setlength{\tabcolsep}{2pt}
\begin{scriptsize}    
\begin{tabular}{c | c| c| c| c| c| c| c}
\hline
Dataset & S-GAN  &{SoPhie} & {PECNet} & Trajectron  & {Trajectron++} & \begin{tabular}{@{}c@{}}TrajPRed \\ w/o R\end{tabular}& TrajPRed \\
\hline
  ETH &  15.70  & 10.54  & 4.18 &  {2.99}  & \textbf{2.98} &{3.63} & 4.15  \\
\hline
Hotel & 8.10 & 5.78  & 0.87 &  2.26  & 0.79 & 0.74 & \textbf{0.56}  \\
\hline
 Univ & 2.88 & 3.70  & 1.61 &  1.05 & 0.83 & 0.89 & \textbf{0.81}  \\
\hline
Zara 1 & 1.36 & 3.03  & 0.62 &  1.86  & \textbf{-0.08}  & 0.16 & {0.07}  \\
\hline
Zara 2 & 0.96 & 3.34 & -0.29 &  0.81 & \textbf{-1.25} & -0.24 & {-0.49}  \\
\hline
Average & 5.80 & 5.28 & 1.40 & 1.79 & \textbf{0.65} & 1.04 & {1.02}  \\
\hline
\end{tabular}
\end{scriptsize}

\end{center}

  \label{Tab: compare_nll}
\end{table}

\begin{table}[ht]
\caption{Robustness study results (ADE/FDE error increase in meters) of our method and other approaches under the perturbation on the ETH-UCY datasets. Lower is better. The lower displacement error increments (error inc) of TrajPRed indicate that region-based relation learning is less susceptible to noise perturbation.}

  \begin{center}
 \setlength{\tabcolsep}{3pt}
\begin{scriptsize}

\begin{tabular}{c|c|c|c|c|c}
\hline
\multirow{ 2}{*}{Dataset} & \multicolumn{5}{c}{ADE/FDE Error Increase (Best of $20$ Predictions)}             \\ 
 & \multicolumn{1}{c|}{S-GAN} &{SoPhie}\cite{sadeghian2019sophie} & {PECNet}\cite{mangalam2020not}   & {Trajectron++}\cite{salzmann2020trajectron++}                & \multicolumn{1}{c}{TrajPRed}                   \\ \hline
ETH    &  0.346/0.540  & 0.320/0.265 & 0.314/0.316 & 0.311/0.284 & \textbf{0.231}/\textbf{0.257} \\ \hline
Hotel  &  0.598/0.908 & 0.525/0.740 & {0.502}/\textbf{0.467} &0.507/0.648  & \textbf{0.425}/{0.473} \\ \hline
Univ   &  0.350/0.522 & 0.315/0.275 & 0.349/\textbf{0.215} &0.362/0.374   & \textbf{0.278}/{0.303} \\ \hline
Zara 1 &  0.484/0.730 & 0.410/0.542 & 0.436/0.449 & 0.478/0.571  & \textbf{0.360}/\textbf{0.436}  \\ \hline
Zara 2 &  0.552/0.872 & 0.402/0.513 & 0.416/0.489 & 0.513/0.617 & \textbf{0.314}/\textbf{0.351} \\ \hline
\end{tabular}
    
\end{scriptsize}
\end{center}

  \label{Tab:robustness_relation_error}
\end{table}

In addition to measuring the optimal behavior using the minimum ADE and FDE, we use the {KDE NLL} metric to evaluate the full prediction distributions. The KDE NLL is a measurement that maintains the generated full distributions and compares the log-likelihood of the ground truth under model prediction distributions. In general, the lower NLL, the closer the model output distribution fits the test data. {We compare the average NLL of our full model, the model with only the multi-goal estimation module (TrajPRed w/o R), and other generative-based methods} in Table~\ref{Tab: compare_nll}.
{Compared to S-GAN and Sophie, our method and PECNet achieve lower NLL, indicating that conditioning on the estimated goal to generate distributions of trajectories can better fit the human behavior of the test data. The average NLL further decreases in most scenes when adding the region-based relation module to our model with only the goal estimation. This observation implies that the predicted distributions can better fit the ground truths along the prediction horizon when incorporating the region-based relation representations. We also find that when the pedestrians are less interactive and their future trajectories are more dependent on individual goals, i.e., the samples in the ETH scene, imposing region-wise relation learning might confuse the distribution prediction. Among all methods, Trajectron++ achieves the overall lowest NLL averaged across all scenes, since it utilizes the physics-driven movement dynamics and optimizes the predicted distribution via log-likelihood maximization.}

\textbf{Robustness of Social Interaction Learning} We study the robustness of the region-based and edge-based relation representations under the perturbation of agents' spatial position states. To simulate the perturbation, we add Gaussian noise with a standard deviation of $0.1$ meters to the observation states. Edge-based spatial relation representations are more distorted since the perturbation is directly applied to the spatial states. Nevertheless, region-based relations are less sensitive to the perturbation due to the utilization of smoothed density as the basis. We compare the displacement error increase of our framework and other frameworks with typical edge-based relation learning on the perturbed test data. As the ADE/FDE error increase in Table~\ref{Tab:robustness_relation_error} shows, utilizing regional density to learn region-wise social relations is less vulnerable to the noise perturbation in the position space.

\begin{figure}[th]

  \centering
  {\includegraphics[width=\columnwidth]{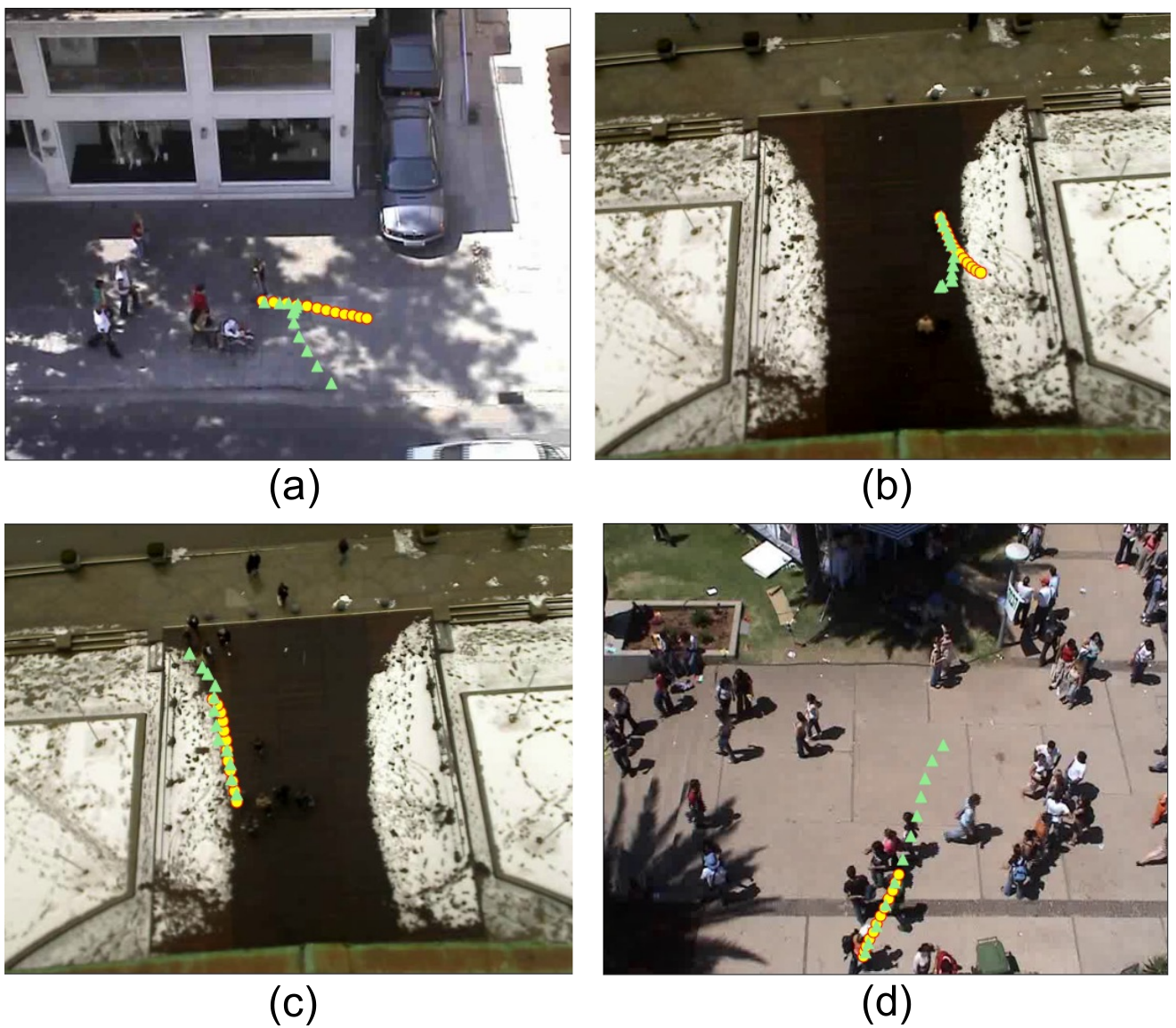}}

\caption{{Examples of prediction errors on ETH-UCY. (a) and (b) illustrate the scenarios of abrupt direction change that our model misses. (c) and (d) show the scenarios of velocity mismatch between our predictions and the ground truths. Green triangles represent the ground truth future trajectories and yellow points represent the associated predicted trajectories.}}
\label{fig:error_analysis}

\end{figure}

{We further evaluate our proposed framework on SDD as shown in Table} ~\ref{Tab:comparative_results_sdd}. {Our method reduces the state-of-the-art prediction errors by $27.61\%$/$18.20\%$ under ADE and FDE metrics, respectively. We reason that the larger improvement in ADE is caused by the advantage of the region-based relation module in capturing social interactions from a global perspective. The comparison demonstrates the efficacy of our method on both ETH/UCY and SDD.}

{\textbf{Prediction Error Analysis} We qualitatively illustrate some examples of prediction errors generated by our method in} Fig.~\ref{fig:error_analysis}. {We show the predictions generated by the estimated optimal goals. (a) and (b) show the scenarios of abrupt direction change in the ground truths that our method misses. Additionally, our framework can be less able to capture the change in the future velocity as shown in (c) and (d), given that the social interaction patterns remain consistent.}

\begin{table*}[th]
\caption{{Trajectory prediction results (ADE/FDE in pixels) of state-of-the-art methods and our approach on Stanford Drone Dataset (SDD). Lower is better.}}

  \begin{center}
\begin{tabular}{c | c| c| c | c| c | c| c}
\hline
Method & S-GAN\cite{gupta2018social} & SoPhie\cite{sadeghian2019sophie} & PECNet\cite{mangalam2020not} & Trajectron++\cite{salzmann2020trajectron++} & CSCNet\cite{xia2022cscnet} &  VDRGCN\cite{su2022trajectory} & TrajPRed \\
\hline

ADE/FDE & 27.23/41.44 & 16.27/29.38 & 9.96/15.88 & 19.30 / 32.70 &  14.63/26.91 & 12.5/17.9 & \textbf{7.21/12.99} \\
\hline
\end{tabular}
\end{center}

  \label{Tab:comparative_results_sdd}
\end{table*}

\subsection{Implementation Details} All models are trained with batch size 64, initial learning rate 0.001, and an exponential learning rate scheduler \cite{li2019exponential} on a single NVIDIA TITAN XP GPU for 100 epochs. We employ the same KL weight annealing strategy as in trajectron++\cite{salzmann2020trajectron++} for fair comparisons. For training the autoencoder, we resize the generated trajectory maps to $80\times80$ with reasonable computation complexity. To prevent over-fitting to the trajectory maps generated using sparsely annotated positions, we augment the maps in each scene. Specifically, we rotate all trajectories in a scene around its origin by $\gamma$, where $\gamma$ ranges from $0$\textdegree\ to $360$\textdegree\ with $30$\textdegree\ intervals.

\section{Conclusion}
In this work, we introduce a trajectory prediction framework that encompasses the modeling of two essential stimuli of human behavior, external social interactions, and individual goals. We propose a robust relation learning paradigm via region-wise temporal dynamics to model the social compliance between agents. The robustness experiments indicate that the region-based relation representations are less vulnerable to spatial noise perturbation compared to approaches with edge-based relation learning. In addition, we present an empirical study of goal estimation to uncover the discrepancy between training and test behavior data, motivating the non-deterministic goal estimation to account for human stochasticity. Specifically, a CVAE is exploited to estimate multiple plausible goals. Finally, we integrate the proposed region-based relation learning module and the multi-goal estimation into our full framework to model social compliance and individual goals. We show that the diverse prediction benefits from the region-based relation module in generating position distributions that better fit the ground truth.  The comparative experiments demonstrate that our framework achieves superior or comparative performance against several state-of-the-art methods on the ETH-UCY dataset and Stanford Drone Dataset.


\appendices


\section*{Acknowledgment}

The authors would like to thank the members of the Omni Lab for Intelligent Visual Engineering and Science (OLIVES) at Georgia Tech, and the reviewers for their feedback. This work was funded by a Ford-Georgia Tech Alliance Project.

\ifCLASSOPTIONcaptionsoff
  \newpage
\fi

\bibliographystyle{IEEEbib}
\bibliography{its}
\begin{IEEEbiography}[{\includegraphics[width=1in,height=1.25in,clip,keepaspectratio]{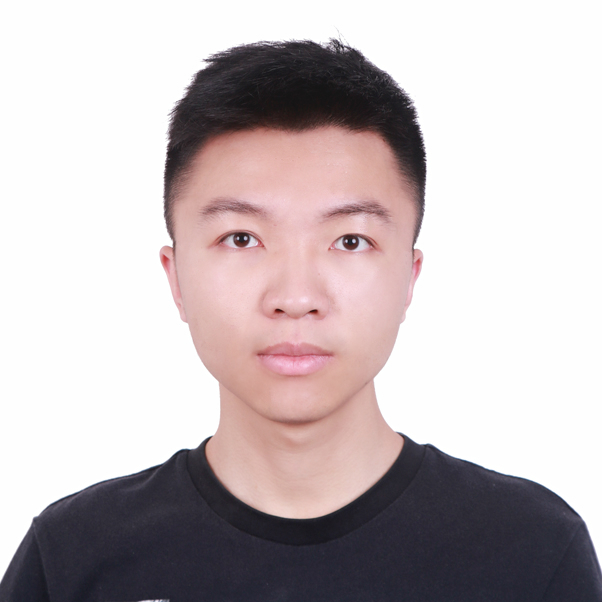}}]{Chen Zhou}
(Student Member, IEEE) received the B.E. degree from the University of Science and Technology Beijing and the M.S. degree from the Georgia Institute of Technology, where he is currently pursuing the Ph.D. degree with the Omni Lab for Intelligent Visual Engineering and Science (OLIVES). He has been working in the fields of machine learning and image and video processing. His research interests include trajectory prediction and learning from label disagreement.
\end{IEEEbiography}
\vspace{-0.5cm}

\begin{IEEEbiography}[{\includegraphics[width=1in,height=1.25in,clip,keepaspectratio]{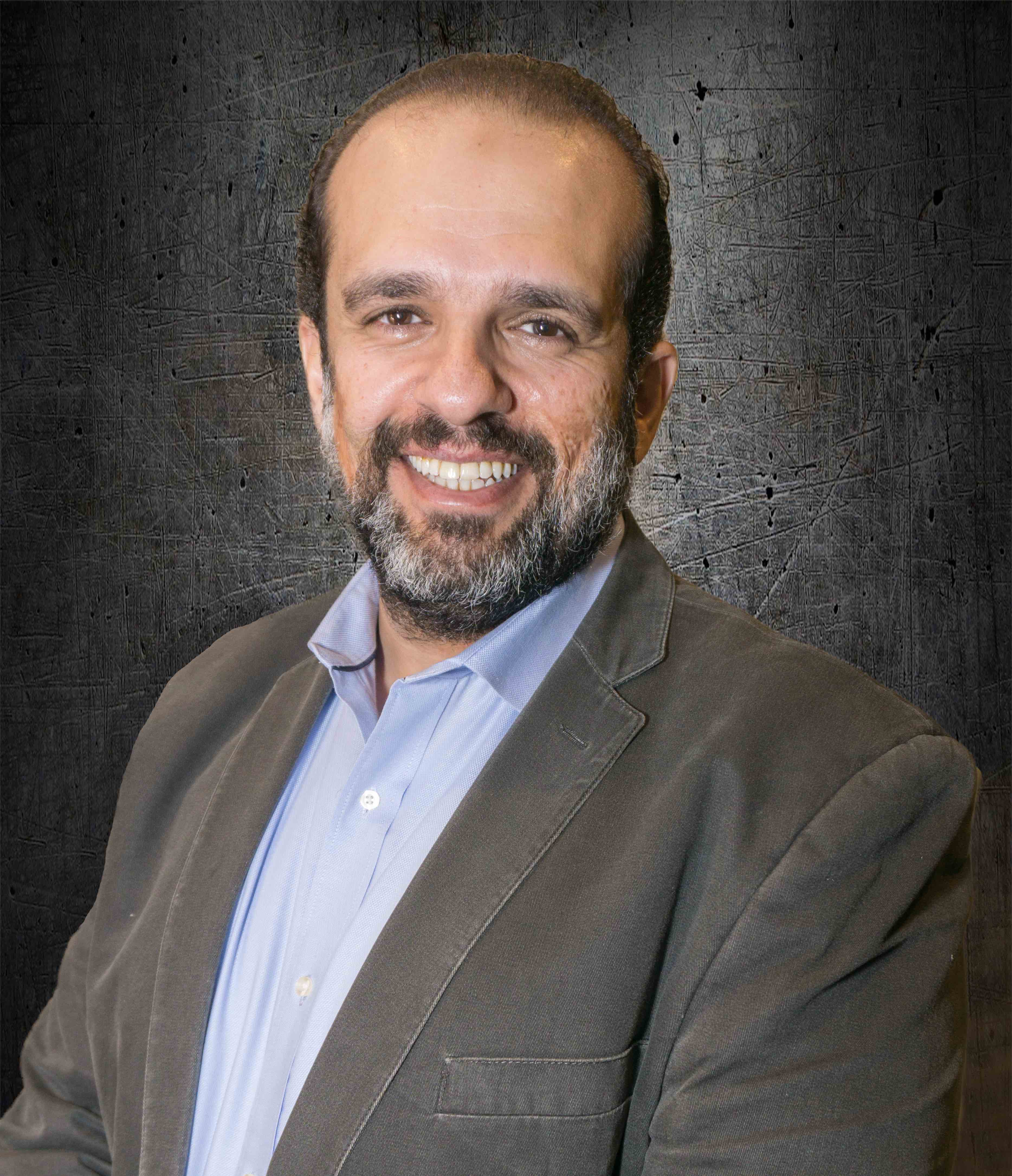}}]{Ghassan AlRegib}
(Fellow, IEEE) is currently the John and Marilu McCarty Chair Professor in the School of Electrical and Computer Engineering at the Georgia Institute of Technology. His research group, the Omni Lab for Intelligent Visual Engineering and Science (OLIVES), works on robust and interpretable machine learning algorithms, uncertainty and trust, and human-in-the-loop algorithms. The group has demonstrated their work on a wide range of applications, such as autonomous systems, medical imaging, and subsurface imaging. He was a recipient of the ECE Outstanding Junior Faculty Member Award in 2008, and the 2017 Denning Faculty Award for Global Engagement. He and his students received the Beat Paper Award in ICIP 2019.  He has participated in several service activities within the IEEE and served on the editorial boards of several journal publications. He served as the Technical Program Co-Chair for ICIP 2020 and ICIP 2024. He served on the editorial boards of IEEE Transactions on Image Processing from 2019 to 2022, and the Elsevier Journal Signal Processing: Image Communications from 2014 to 2022. He served as an Area Editor for Columns and Forums in IEEE Signal Processing Magazine from 2009 to 2012. He led a team that organized the inaugural IEEE VIP Cup in 2017, and the IEEE VIP Cup in 2023. He has been a witness expert in several patent infringement cases.
\end{IEEEbiography}
\vspace{-0.5cm}

\begin{IEEEbiography}[{\includegraphics[width=1in,height=1.25in,clip,keepaspectratio]{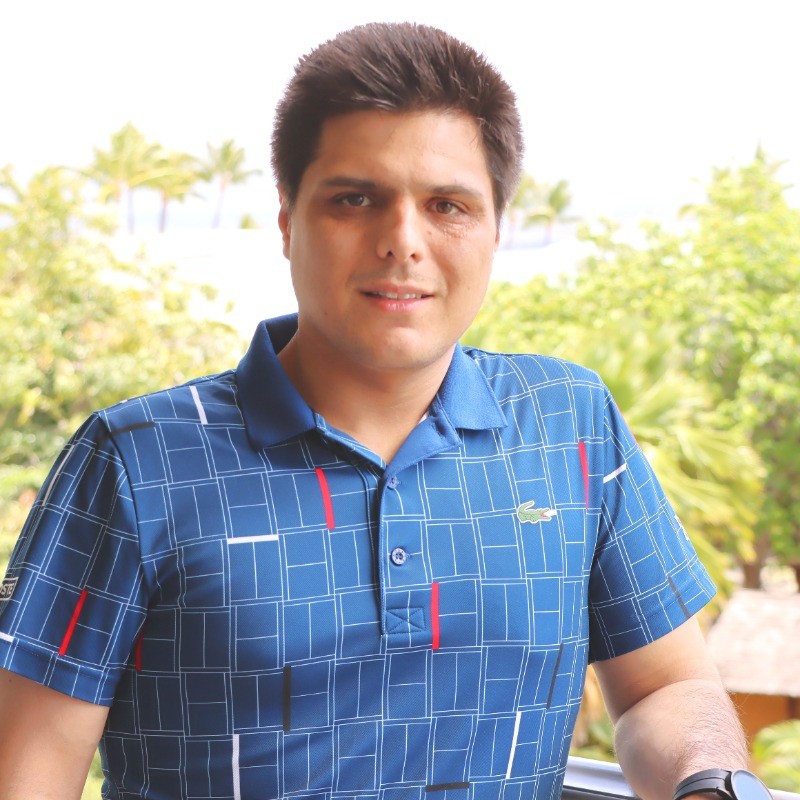}}]{Armin Parchami}
received the B.E. degree in Software Engineering and the M.Sc. degree in Artificial Intelligence from Bu-Ali Sina University and the Ph.D. degree in Computer Science from UTA in 2017. His dissertation focused on single-shot face recognition using deep learning algorithms for security applications. Previously, he contributed significantly to the field of autonomous vehicles during his tenure at Ford, where he held the position of Director of Perception, leading the development of perception algorithms for L2+ autonomy. Recently, he transitioned to Snorkel AI as a Principal ML Research Scientist, focusing on programmatic labeling solutions for computer vision tasks, continuing to drive innovation in machine learning and artificial intelligence.
\end{IEEEbiography}
\vspace{-0.5cm}

\begin{IEEEbiography}[{\includegraphics[width=1in,height=1.25in,clip,keepaspectratio]{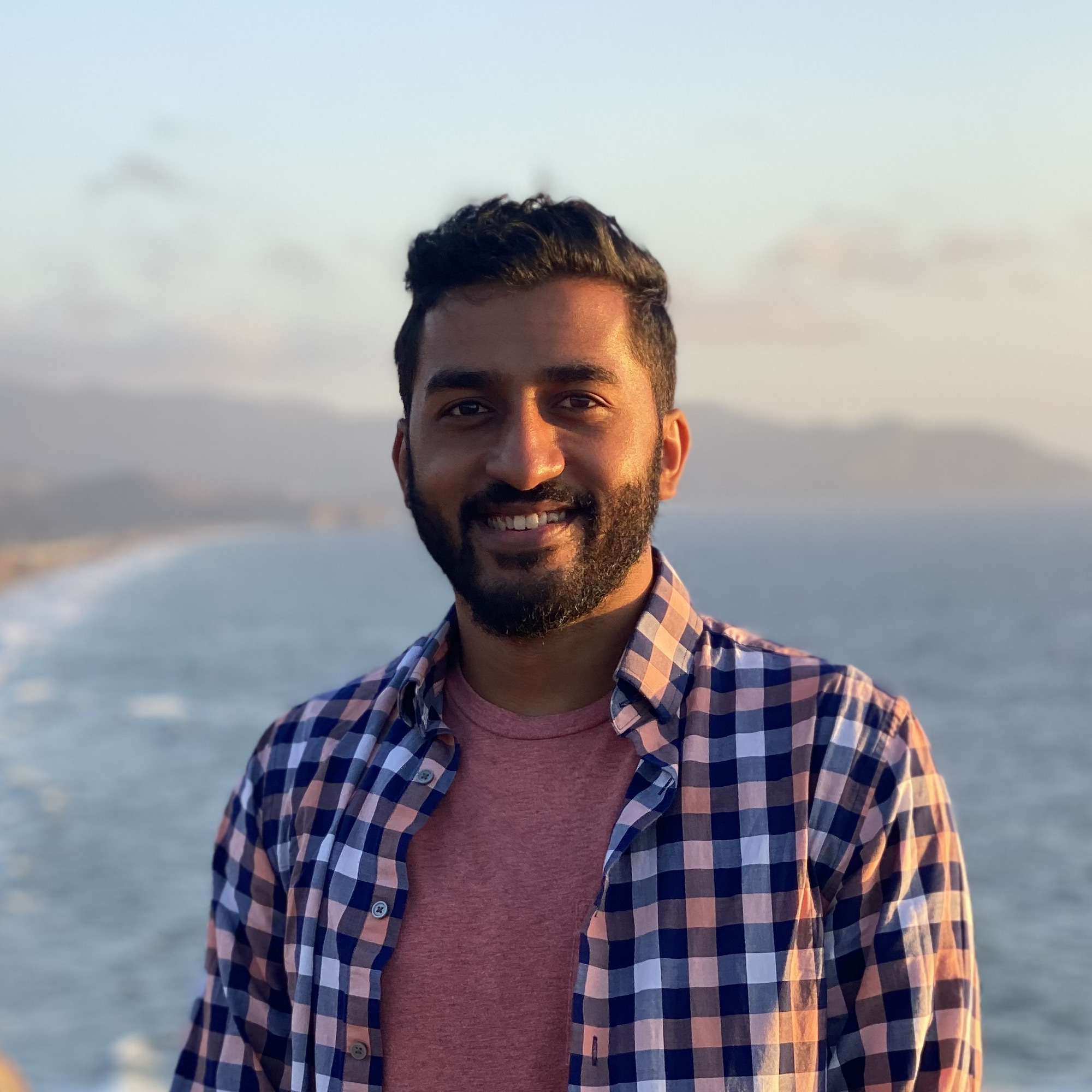}}]{Kunjan Singh}
received the Bachelor's degree in Computer Engineering from the University of Michigan and the Master's degree in Computer Science with a specialization in Machine Learning from Georgia Tech in 2023. During his time at Ford, he has made valuable contributions to the development of computer vision solutions for autonomous vehicles, smart infrastructure technology, and ADAS.
\end{IEEEbiography}
\vspace{-0.5cm}



\end{document}